\documentclass[times,twocolumn,final,authoryear]{elsarticle}

\usepackage{ycviu}
\usepackage{framed,multirow}

\usepackage{amssymb}
\usepackage{latexsym}

\usepackage{url}
\usepackage{xcolor}
\definecolor{newcolor}{rgb}{.8,.349,.1}

\usepackage{microtype}
\usepackage{graphicx}
\usepackage{subfigure}
\usepackage{algorithm}
\usepackage{algorithmic}
\usepackage{tikz}

\usepackage{mathtools}
\usepackage{amsthm}

\usepackage{fancyhdr}

\fancypagestyle{Whatever}{
\fancyhf{} 
\fancyfoot[C]{ \color{red}{This paper is under consideration at Computer Vision and Image Understanding.} Page \thepage} 
}

\journal{Computer Vision and Image Understanding}

\begin{document}

\ifpreprint
  \setcounter{page}{1}
\else
  \setcounter{page}{1}
\fi

\begin{frontmatter}

\title{ParticleAugment: Sampling-based Data Augmentation}

\author[1]{Alexander \snm{Tsaregorodtsev}\corref{cor1}} 
\cortext[cor1]{Corresponding author: 
  Tel.: +49-731-50-27030;}
\ead{alexander.tsaregorodtsev@uni-ulm.de}
\author[1]{Vasileios \snm{Belagiannis}}

\address[1]{Ulm University, Albert-Einstein-Allee 41, 89081 Ulm, Germany}

\received{1 May 2013}
\finalform{10 May 2013}
\accepted{13 May 2013}
\availableonline{15 May 2013}
\communicated{S. Sarkar}

\begin{abstract}
We present an automated data augmentation approach for image classification. We formulate the problem as Monte Carlo sampling where our goal is to approximate the optimal augmentation policies. We propose a particle filtering scheme for the policy search where the probability of applying a set of augmentation operations forms the state of the filter. We measure the policy performance based on the loss function difference between a reference and the actual model, which we afterwards use to re-weight the particles and finally update the policy. In our experiments, we show that our formulation for automated augmentation reaches promising results on CIFAR-10, CIFAR-100, and ImageNet datasets using the standard network architectures for this problem. By comparing with the related work, our method reaches a balance between the computational cost of policy search and the model performance. Our code will be made publicly available. 
\end{abstract}

\begin{keyword}
\MSC 41A05\sep 41A10\sep 65D05\sep 65D17
\KWD Keyword1\sep Keyword2\sep Keyword3

\end{keyword}

\end{frontmatter}


\section{Introduction}

Data augmentation traditionally improves the generalization of deep neural networks. It increases the training set size, as well as data diversity and thus prevents over-fitting by acting as a regularizer. It is common in computer vision tasks such as image classification~\citep{imagenet_cvpr09, he2015deep}, object detection~\citep{munjal2020joint, ren2016faster}, and semantic segmentation~\citep{chen2017deeplab, dawoud2020few}. Although hand-designed image augmentations, like translations, rotations, or flips, work well in practice, there is a recent family of methods that shows superior performance by automating data augmentation.

Automated image augmentation algorithms look for the optimal set of augmentations that minimize the objective function of the neural network. It is common to formulate the optimal augmentation policy as a sequence of individual augmentation operations~\citep{cubuk2019autoaugment, lim2019fast, ho2019population, cubuk2019randaugment}. Each operation, e.g.~ rotation, can be parametrized by the application probability and augmentation magnitude~\citep{cubuk2019autoaugment}. Finding the optimal set of augmentations is challenging because of the high dimensionality of the problem. For instance, there are usually more than ten available image augmentations where the frequency and augmentation order have to be set as well. Furthermore, it is necessary to constantly monitor neural network performance during training with augmentation policies, which is computationally expensive.

AutoAugment~\citep{cubuk2019autoaugment} is one of the first successful attempts to automatize data augmentation. It relies on reinforcement learning to determine the optimal augmentation policies to apply during training. One of its drawbacks is its massive computational demand, as during optimization a smaller proxy model, which is similar to the target model, needs to be retrained multiple times. This limitation has been partially solved by the Fast AutoAugment algorithm~\citep{lim2019fast} at the cost of model performance. The computational complexity of policy search has been also addressed by the Population-Based Augmentation (PBA)~\citep{ho2019population} algorithm, knowledge distillation~\citep{wei2020circumventing}, and RandAugment~\citep{cubuk2019randaugment}. Nevertheless, these approaches either rely on heuristics~\citep{cubuk2019randaugment} to relax the computation requirements or sacrifice model performance~\citep{lim2019fast}. In this work, we find a balance between the computational cost of policy search and the model performance (see Sec.~\ref{lbl:ablation}).

We present ParticleAugment, an algorithm to learn dynamic augmentation policies for image classification tasks. Based on Monte Carlo sampling, we explore a dynamic policy search space. In detail, by defining the policy space as a distribution function, we efficiently seek for the optimal policies with a particle filtering scheme. The probability of applying a set of augmentation operations forms the state of the filter, while a simple constant position model is responsible for the filter state transition. The algorithm starts with randomly initialized policies which are necessary for the first model training. After some training epochs, the policy performance is measured by comparing the loss difference between our policy-trained model and reference model. The measured policy performance is then used to re-weight the filter particles and update the policies. Our approach is illustrated in Algorithm~\ref{alg:example}. More importantly, we do not train multiple neural network models for the prediction step of each augmentation particle. Instead, we rely on training a single model, which is updated while using all augmentation particles. Also, as the particle filter is constantly updated during the training procedure we can continuously optimize policies, making them dynamic. Our approach only adds a small overhead to the model training process. In our evaluations, we reach state-of-the-art classification accuracy on standard benchmarks such as CIFAR-10, CIFAR-100, and ImageNet compared to prior works. We will make our implementation publicly available upon acceptance and provide all necessary code and hyper-parameters to reproduce our results.

\section{Related Work}

\paragraph{Manual image augmentation}

Image augmentation is a standard pre-processing step for visual tasks such as image classification~\citep{he2015deep} and object detection~\citep{ren2016faster}. Spatial transformations, such as flipping, cropping, translation, rotation, scaling as well as color transformations are normally used for training deep neural networks, e.g.~~ResNet~\citep{he2015deep}, Faster-RCNN~\citep{ren2016faster}, and YOLO~\citep{redmon2016look, redmon2018yolov3}. The motivation of pre-processing is to diversify and enrich the training set, thus making the network generalize better and prevent over-fitting. For example, Cutout~\citep{devries2017improved} showed impressive generalization improvement while being a cost-effective augmentation method. It removes entire image patches and therefore pushes the sample out of the source distribution. Two similar techniques are Thumbnail~\citep{xie2021thumbnail}, Mixup~\citep{zhang2017mixup}, and CutMix~\citep{yun2019cutmix}, which fill the cutouts with patches from the same image or other image samples and merge their labels according to their patch size. While the aforementioned augmentations rely on standard pixel-wise and geometric operations, another family of methods performs the augmentations with deep neural networks. For instance, Generative Adversarial Networks (GANs)~\citep{ratner2017learning} modify an image or even generate new data~\citep{antoniou2018data, sixt2017rendergan, zhu2017data}. However, the prior work~\citep{cubuk2019autoaugment, ho2019population} shows that carefully selecting the augmentation types can improve generalization performance compared to manual augmentation.

\paragraph{Automated image augmentation}
Unlike the predefined transformations, automated image augmentation approaches aim to find the set of transformations that maximize the performance for specific network architecture and dataset. AutoAugment~\citep{cubuk2019autoaugment} is one of the first approaches to automatize image augmentations. It is based on reinforcement learning to find the optimal augmentation policies, i.e. transformations, with the policy proposal network. While AutoAugment is still one of the best performing algorithms, its biggest drawback is the computational complexity.Every iteration of AutoAugment needs a full training of small proxy networks and contains about 30 optimization parameters, resulting in enormous GPU hour requirements. To lower these requirements, Fast AutoAugment~\citep{lim2019fast} relies on density matching and the Tree-Parzens estimator algorithm (TPE)~\citep{10.5555/2986459.2986743, 10.5555/3042817.3042832} to achieve similar performance to AutoAugment. OHL-AutoAugment~\citep{lin2019online}, Adversarial AutoAugment~\citep{zhang2020adversarial}, and Augmentation-Wise Weight Sharing (AWS)~\citep{tian2020aws} are also variants of AutoAugment, which further improve performance of trained networks while reducing the computational power required for training.
Population-Based Augmentation (PBA)~\citep{ho2019population} explores a different approach to speed up AutoAugment. A population of networks is trained with various augmentations. Then, the best-performing network is chosen and perturbed to obtain a population of networks. This, however, means that for each training epoch, multiple dozens of networks need to be trained, which, while being faster than AutoAugment, still consumes a lot of training time. Faster AutoAugment~\citep{hataya2020faster} and Differentiable Automatic Data Augmentation (DADA)~\citep{li2020differentiable} are a different class of algorithms with reduced computational cost based on a differentiable policy search, similar to neural architecture search. RandAugment (RA)~\citep{cubuk2019randaugment} is a leading algorithm that requires only two hyper-parameters $n$ and $m$ as well as a set of augmentation to choose from. RA then works by sampling $n$ augmentations of magnitude $m$ from the chosen set and applying them on the data. To achieve good performance, both parameters are optimized for each dataset and network architecture. The limited amount of hyper-parameters and small computational overhead makes RandAugment work well in practice. However, it does not consider any time dependencies that may be present during a training procedure. Currently, there is not any automatic augmentation method that performs optimal augmentation policy search for each training pass with a small computational overhead. Furthermore, most methods require multiple proxy models and a massive number of computations outside of the regular training.
We propose a new automatic augmentation method based on Monte Carlo sampling to address these issues.

\paragraph{Monte Carlo Sampling}

We derive motivation from control theory to formulate our problem. In control theory, a common task is to estimate the system states from noisy measurements for implementing feedback and feed-forward control. A popular approach to solve this task is the particle filter. It was first introduced as Sequential Importance Resampling (SIR) filter in~\citep{Kitagawa1996} and Bootstrap filter in~\citep{Gordon1993}. The SIR filter itself is a part of a broader set of methods, called Monte Carlo Methods (MCM) \citep{bronshtein2013handbook}.
MCM is particularly suitable for the task of finding augmentation policies, as finding policies is not a low-dimensional optimization problem, e.g., 15 distinct parameters. In this work, we explore the SIR filter approach to estimate the augmentations and optimize for the optimal policy schedules, instead of estimating fixed policies.

\section{Method}

\newcommand{\comment}[1]{}

Let $\mathcal{F} =\{f_{k}(\nu,\ p_{k})\}_{k=1}^{n}$ be a set of $n$ functions, each of which represents one augmentation operation, applied to the data sample $\nu$. In our problem, the data sample corresponds to the image $\nu \in \mathbb{R}^{H\times \mathrm{W} \times \mathrm{3}}$ (where $H$ and $W$ the image height and width) with the one-hot vector label $\mathbf{y} \in \{0,1\}^{C}$, such that $\sum_{c=1}^{C}\mathbf{y}(c)=1$ for a C-category classification problem. Furthermore, the augmentation function $f_{k}(\cdot)$ is parametrized by the probability  of applying the augmentation $p_{k} \in [0,\ 1]$. While every augmentation can also be parametrized by its magnitude $m$, we use static one global magnitude in order to minimize the problem complexity. The $\mathcal{F}$ augmentations are sequentially applied to the image $\nu$ based on $p_{k}$. Given the training set $\mathcal{D}=\{(\nu_d,\mathbf{y}_d)\}_{d=1}^{|\mathcal{D}|}$ with images $\nu_d$ and labels $\mathbf{y}_d$, our goal is to train a deep neural network with optimal augmentations. To that end, we seek the optimal policy schedules $\mathbf{x}_{i,t} = [p_{1,t}\ \hdots \ p_{n,t}]$ for every training epoch $t$, where $i\in{1, \hdots, r}$ is the policy index and $r$ is the number of policies. We propose to learn the augmentation policies with Monte Carlo sampling where we rely on a particle filter to obtain the policy $\mathbf{x}_{i,t}$. Below, we present the particle filter background and our formulation for augmentation policy learning.

\subsection{Particle Filter}

Particle filters are Monte Carlo sampling algorithms that use state and measurement models to estimate system states from noisy observations~\citep{Gordon1993, Kitagawa1996}. Our algorithm is based on the Sequential Importance Resampling (SIR) approach~\citep{Kitagawa1996}. Consider the following equations:
\begin{equation}
    \mathbf{S} = \left[ (\mathbf{x}_{i,t}, w_{i,t}) \right]^{r}_{i=1},
\end{equation}
\begin{equation}
    \mathbf{x}_{i,t+1} = f(\mathbf{x}_{i,t}) + \mathbf{v}_{i,t},
    \label{eqn:pmodel}
\end{equation}
\begin{equation}
    w_{t+1} = \frac{\eta \cdot g(\mathbf{z}_{t+1} - \mathbf{x}_{t+1}) \cdot w_{t}}{\sum_{i=0}^r \eta \cdot g(\mathbf{z}_{t+1} - \mathbf{x}_{i,t+1}) \cdot w_{i,t}}\,
    \label{eqn:imodel}
\end{equation}
\begin{equation}
    \mathbb{E}[\mathbf{x}_{t+1}] = \sum_{i=1}^{r}w_{i,t+1}\mathbf{x}_{i,t+1}\ . \label{eqn:fs}
\end{equation}
These equations describe the implementation of a simple SIR filter. Here $\mathbf{S}$ describes the filter state consisting of $r$ particles, where each particle is described by its state $\mathbf{x}_{i,t}$ at timestamp $t$ and its associated weight $w_{i,t}$. The particle weight can be interpreted as the probability that its state coincides with the actual system state. To update the filter states, a system model described by a transition function $f$
and sampled model (process) noise $\mathbf{v}_{i,t}$ is applied to calculate a prediction in Eqn.~\ref{eqn:pmodel}. This prediction is then compared to a state measurement $\mathbf{z}_{i,t+1})$ by a mapping g (Eqn.~\ref{eqn:imodel}), which is normally chosen in such a manner that the resulting updated weight accurately reflects the probability of its associated state, factoring in measurement errors. Finally, to estimate a single output state, the first moment (Eqn.~\ref{eqn:fs}) as well as higher moments can be calculated over all states. To combat degeneration effects, where one weight tends to 1 while all other tend to 0, particle resampling can be applied. To this end, new particles are drawn from current particles according to their associated weights. Resampling therefore helps with the removal of particles which do not represent the current system state adequately.

In our problem, we start with a static, augmentation model, which translates to a constant model $f(\mathbf{x})=\mathbf{x}$. In this case, a Gaussian process noise is assumed, which is common when there is not prior information on the underlying model. The relative loss difference of a model (Eqn.~\ref{eqn:val2}) is used as a measurement for the innovation step.
With these assumptions, we can construct an SIR filter performing online augmentation policy optimization, as explained below.

\subsection{Augmentation Policy Learning}
\label{lbl:apl}

We seek to find the optimal augmentation policies. To this end, we define a single augmentation policy at the training epoch $t$ as a particle with the state and associated weight:
\begin{equation}
    \mathbf{x}_{i,t} = [ p_1,\ p_j,\ \hdots,\ p_n ], w_{i,t}\in[0,1],\ i\in \{1, \dots, \ r\}\ .
\end{equation}

In this formulation, $n$ represents the number of different applicable augmentations, e.g.~$n=15$ if we rely on the same setup as RandAugment~\citep{cubuk2019randaugment}, $r$ is the particle number and $i$ the particle index. The second index $t$ denotes the current training epoch of the filter. The policy element $p_j$ represents the application probability of augmentation $j$ and is confined in the range $[0, 1]$. We can thus have a variable number of augmentations from $0$ to $n$ being applied during every policy invocation, as every $p_j$ is used to independently sample or not sample each of the $n$ available augmentations, resulting in a dynamic policy scheme. Moreover, it is worth noting that the augmentations are applied sequentially. The order of augmentation does not play an important role in the outcome. We empirically prove it in our evaluations, as shown in Sec.~\ref{lbl:ablation}. Overall, each particle is a distinct augmentation policy that is re-weighted during every filter step to optimize the overall augmentation strategy. Furthermore, each policy is changed based on the state transition model.

\paragraph{State transition model}
\label{subs:stm}

We rely on the constant position and constant velocity models~\citep{cvmodel} to define the state transition model for augmentation policies. In our evaluations, we observed that both models work well in practice regardless of their simple formulation. A Gaussian distribution $\mathcal{N}(0, \sigma^{2})$ with zero mean is assumed for sampling transitions, as we have no prior knowledge of the actual noise process. The predicted state is given by:
\begin{equation}
     \mathbf{x}_{i,t+1} = \mathbf{x}_{i,t} - \mathbf{c} + \mathcal{N}(0, \sigma^{2})\ , \label{eqn:pnoise}
\end{equation}
where $\mathbf{c}$ represents constant velocity, which we use to actively bias the augmentation probabilities towards a desired direction. We empirically observe that the constant velocity can further improve model performance in some situations (see experiments in Sec.~\ref{lbl:smallscale}). If any elements of $\mathbf{x}_{i,t+1}$ are not in the range of $[0,\ 1]$ after the state transition, they are clipped to fit the interval $[0,1]$ to represent valid probabilities. The constant velocity model can be changed to the constant position model by setting the velocity $\mathbf{c}$ to a zero vector. In Sec.~\ref{lbl:ablation}, we present an ablation study with both models to show the possible velocity impact.

\paragraph{Filter prediction step}

The prediction step consists of first cloning the trained model $\mathcal{M}_{\mathrm{ref},t}$ of the current training epoch $t$ to $\mathcal{M}_{\mathrm{upd,t}}$. The cloned model $\mathcal{M}_{\mathrm{upd},t}$ is trained for another $e_p$ epochs with a random subset of the training set $\mathcal{D}$ with the same label distribution called $\mathcal{D}_{tp}$ to reduce the training time. During training of $\mathcal{M}_{\mathrm{upd},t}$ we rely on all augmentation policies, as obtained from Eq.~\ref{eqn:pnoise}. In particular, during the training of $\mathcal{M}_{\mathrm{upd},t}$, each data sample is augmented with only one augmentation policy $\mathbf{x}_{i,t+1}$. One novelty of this training procedure is that the policy for each sample is randomly chosen from all particles and the probability of a specific policy being chosen is determined by its filter weight. In this case the filter weights $w_{i,t}$ represent a discrete probability distribution. Therefore we randomly apply \textit{all} particles based on their weights while training the \textit{same} model, instead of training a separate model for each policy, thus saving computational resources. We motivate this approach by comparing it to data mini-batching in gradient-based learning. In the mini-batch stochastic gradient descent, mini-batching is performed to significantly increase training speed and suppress the gradients noise obtained from single samples. Our case is similar, as we increase speed by only training one single network for one training epoch, but we still see the impact of every single policy on the model in an averaged manner, similar to gradients during mini-batching.

\paragraph{Measuring policy performance}

We propose a scheme that utilizes the training loss function to obtain a quantitative performance measurement for each augmentation policy. After training $\mathcal{M}_{\mathrm{upd},t}$, the original training set $\mathcal{D}$ is split again with a stratified shuffle split strategy to obtain a measurement (validation) subset $\mathcal{D}_{vp}$. This measurement subset is then used to benchmark the effect of each specific augmentation policy. In particular, the subset is augmented with a policy $\mathbf{x}_{i,t+1}$ to obtain augmented samples $\mu_k$ and then fed to both $\mathcal{M}_{\mathrm{ref},t}$ and $\mathcal{M}_{\mathrm{upd},t}$ to calculate their loss difference as:

\begin{equation}
    d_{i,t+1} = \sum_{k=1}^{n_{\mathrm{samples}}}(\mathcal{L}_{\mathcal{M}_{\mathrm{ref},t}}(\mu_k) - \mathcal{L}_{\mathcal{M}_{\mathrm{upd},t}}(\mu_k))
    \label{eqn:val1}
\end{equation}
\begin{equation}
    \delta_{i,t+1} = \frac{d_{i,t+1}}{d_{0,t+1}}\ ,
    \label{eqn:val2}
\end{equation}
where $n_{\mathrm{samples}}$ is the size of the measurement subset $\mathcal{D}_{vp}$ and and $k$ is the index of the sample. We define as $d_{0,t+1}$ the loss difference for non-augmented samples from the measurement subset. It is needed to convert the absolute loss difference $d_{i,t+1}$ to the \textit{relative loss difference} $\delta_{i,t+1}$.
Based on $\delta_{i,t+1}$, we measure the loss improvement when training with the current filter particles. In particular, if we obtain a smaller value after training with $\mathcal{M}_{\mathrm{upd},t}$ using the current particles and augmenting the measurement subset with a specific policy particle, we can infer that the network can handle the current policy better. While if we have a larger $\delta_{i,t+1}$ value for the updated model, then the policy is not helpful for training. Using $\delta_{i,t+1}$, the particle weights can be updated using our empirically derived weight update rule, defined as: \begin{equation}
    \hat{w}_{i,t+1} = (\tanh(\delta_{i,t+1} -1) + 1)^\eta \cdot w_{i,t}\ , \label{eqn:wupd}
\end{equation}
\begin{equation}
    w_{i,t+1} = \frac{\hat{w}_{i,t+1}}{\sum_{j=1}^n \hat{w}_{j,t+1}}\ . \label{eqn:wupd2}
\end{equation}
The denominator of Eq.~\ref{eqn:wupd2} normalizes the weights so that they represent a discrete probability distribution. Also, the $\tanh$ function of Eq.~\ref{eqn:wupd} compresses the arbitrary range of $\delta_{i,t+1}$ to the interval $[0,2]$. In this way, we constrain the weights to positives values and prevent them from degeneration because of high $\delta_{i,t+1}$ values. The update rate $\eta$ is applied as an exponent to compress or expand the update range of the coefficients. Eq.~\ref{eqn:wupd} was designed with the requirement that the relative improvement $\delta_{i,t+1}$ of 1 is mapped to to the original weight. This means no improvement but also no degradation. All $\delta_{i,t+1} > 1$ are mapped to larger weights, while all $\delta_{i,t+1} < 1$ are mapped to smaller weights. After measuring the policy performance, an optional re-sample step is performed if required.
A big difference to the classic SIR filter, though, is that it often uses Eq.~\ref{eqn:fs} to output a single estimated state, whereas we do not aggregate any results and sample all possible particles according to their weights instead. Therefore, we make use of the entire particle filter distribution.

\subsection{Complete Training Algorithm}
\label{lbl:cta}

The full algorithm is described in Alg.~\ref{alg:example}. In the first filter step, the particles are initialized as vectors of length 15 (as we have 15 base augmentations available\footnote{These are Identity, AutoContrast, Equalize, Rotate, Solarize, SolarizeAdd, Color, Contrast, Brightness, Sharpness, ShearX, ShearY, TranslateX, TranslateY, Posterize to match the augmentations available for other methods, while SolarizeAdd was additionally provided by the RA implementation that was used for ParticleAugment.}) with $l$ random elements of each particle set to $0.25$, where $l$ can be interpreted as the maximal amount of augmentations applied by each policy. In Alg.~\ref{alg:example}, we call this method sparseInit(). After the first epoch, the SIR filter is invoked at every $i_f$ epochs. Each invocation consists of the prediction step (Eq.~\ref{eqn:pnoise}), measuring policy performance (Eq.~\ref{eqn:val1}, Eq.~\ref{eqn:val2}), update (Eq.~\ref{eqn:wupd}, \ref{eqn:wupd2}), and the optional re-sample step. The resulting particles serve as the augmentation policies during the next $i_f$ training epochs. During training, each policy $\mathbf{x}_{i,t}$ is sampled independently based on its sampling probability $w_{i,t}$. To simplify our augmentation space, we have the same augmentation magnitude for all operations.

We avoid to initialize with completely random policies because of the conclusions from related algorithms~\citep{lim2019fast, cubuk2019randaugment}. It is shown that applying only a couple of augmentations is enough for most network architectures. Therefore initializing all states of an augmentation policy to a non-zero value will lead to policies that apply too many augmentations and lower the data quality to a degree that neither a neural network nor a human can infer a correct label. This is also addressed in research on knowledge distillation~\citep{wei2020circumventing} and self-training~\citep{zoph2020rethinking}. For this reason, we also apply a sparse initialization.

\begin{algorithm}[tb]
   \caption{ParticleAugment algorithm description. Here, the augment function will augment using policy $\mathbf{x}_{i}$, if no $\mathbf{w}$ is specified. Otherwise, it will augment using a sampled policy $\mathbf{x}_{j}$, where $j$ is sampled from $\mathbf{x}$ according to the weights $\mathbf{w}$.}
   \label{alg:example}
\begin{algorithmic}
   \STATE $\mathbf{x}_{i,0} \leftarrow$ sparseInit() \textbf{foreach} $i\in 1 \hdots r$
   \FOR{epoch $e$ in $[1,\ e_\mathrm{max}]$}
   \STATE $\mathcal{M}_{\mathrm{ref},t} \leftarrow$ train$(\mathcal{M}_{\mathrm{ref},t-1}, \mathcal{D}_\mathrm{train, aug})$
   \IF{$e\ \mathrm{mod}\ i_f = 0$}
   \STATE $\mathbf{x}_{i, t+1} \leftarrow \mathrm{addNoise}(\mathbf{x}_{i,t})$
   \STATE $\mathcal{D}_{tp,i} \leftarrow$ augment$(\mathcal{D}_{tp}, \mathbf{x}_{t+1}, \mathbf{w}_t)$
   \STATE $\mathcal{M}_{\mathrm{upd},t} \leftarrow$ train$(\mathrm{copy}(\mathcal{M}_{\mathrm{ref},t}), \mathcal{D}_{tp,i})$
   \FOR{$i\ \in\ 1 \hdots r$}
   \STATE $\mathcal{D}_{vp,i} \leftarrow$ augment$(\mathcal{D}_{vp},\ \mathbf{x}_{i, t+1})$

   \STATE $\delta_{i,t+1} \leftarrow$ evaluate($\mathcal{M}_{\mathrm{upd},t}, \mathcal{D}_{vp,i}$)
   \STATE $\hat{w}_{i,t+1} \leftarrow \mathrm{updateWeight}(\hat{w}_{i,t}, \delta_{i,t+1})$
   \ENDFOR
   \STATE $\mathbf{w}_{t+1} \leftarrow \mathrm{normalize}(\hat{w}_{1,t+1},\ \hdots,\ \hat{w}_{r,t+1})$
   \IF{$\mathit{N}_{\mathrm{eff}} = \frac{1}{\sum_{j=1}^\mathit{r}w_{i, t}^2} < r \cdot \alpha$}
   \STATE $\mathbf{x}_{i, t+1}, \mathbf{w}_{t+1} \leftarrow \ $ resample($\mathbf{x}_{i, t+1}, {w}_{t+1}$)
   \ENDIF
   \ENDIF
   \ENDFOR
\end{algorithmic}
\end{algorithm}

\section{Experiments}
\label{section:experiments}

We evaluate our approach for image classification, where we report results on the CIFAR-10/CIFAR-100~\citep{cifar10} and ImageNet~\citep{imagenet_cvpr09} datasets. Moreover, we consider the WideResNet~\citep{zagoruyko2017wide}, ResNet~\citep{he2015deep}, and ShakeShake~\citep{gastaldi2017shakeshake} architectures, which are common for this evaluation protocol. We compare our approach with Fast Autoaugment~\citep{lim2019fast}, RandAugment~\citep{cubuk2019randaugment}, DADA \citep{li2020differentiable}, and AutoAugment~\citep{cubuk2019autoaugment}. We rely on the RandAugment augmentation set\footnote{https://github.com/rwightman/pytorch-image-models/blob/master/timm/data/auto\_augment.py (licensed under Apache License 2.0)} for all of our experiments. Next, we discuss the experimental setup and then present our results, as well as a number of ablations studies.

\subsection{Common parameters and setup}

To set all hyper-parameters of our method, we make use of the WideResNet-28-2 model and the CIFAR-10 dataset. Afterwards, we perform all evaluations with the same parameters, unless noted otherwise. For the number of particles $r$, we conclude that 50 particles are sufficient, as can be seen in Table \ref{tbl:abl}. The process noise $\sigma$ is set to 0.05 as we need sufficient variation of the policies. The filter learning rate $\eta$ from Eq.~\ref{eqn:wupd} is set to 1.0. The filter training epoch $e_p$ was set to 1.

The training subset $\mathcal{D}_{tp}$ is around 25000 samples or around 50\% of the original set for CIFAR training and 180000 samples or 15\% for ImageNet, using a stratified shuffle split strategy which preserves the original label distribution. The measurement subset $\mathcal{D}_{vp}$ is chosen to be between 500 and 5000 samples of the training subset $\mathcal{D}$ to limit the computational power needed for the measurement step. We used a filter warm-up delay of 1 epoch, i.e.~the first filter step is performed after 1 epoch, and a filter step $i_f$ of 1 epoch. All experiments were performed with up to 4 NVIDIA 2080Ti GPUs for the ImageNet experiment and only 2 GPUs for all other experiments.

\begin{table}[h]
\begin{center}
\caption{We report the accuracy on the dataset test split after training with ParticleAugment on CIFAR-10/100~\citep{cifar10} and ImageNet~\citep{imagenet_cvpr09} image classification tasks. We compare against AutoAugment (AA)~\citep{cubuk2019autoaugment}, Fast AutoAugment (FAA)~\citep{lim2019fast}, DADA~\citep{li2020differentiable}, and RandAugment (RA)~\citep{cubuk2019randaugment}. The results of these algorithms were taken from their respective papers, dashed lines indicate results that were not reported. Accuracy values are reported in percentage.}
\vskip 0.1in
\resizebox{0.925\textwidth}{!}{\begin{minipage}{\textwidth}
\begin{tabular}{lccccc}
\hline
Dataset & AA & FAA & RA & DADA & PA (Ours) \\
\hline\hline
\textbf{CIFAR-10} & & & & & \\
WideResNet-28-2 & 95.9 & - & 95.8 & - & \textbf{96.4} \\
WideResNet-28-10 & 97.4 & 97.3 & 97.3 & 97.3 & \textbf{97.8} \\
ShakeShake-26x96 & 98.0 & 98.0 & 98.0 & 98.0 & \textbf{98.2} \\
\hline
\textbf{CIFAR-100} & & & & & \\
WideResNet-28-2 & 78.5 & - & 78.3 & - & \textbf{78.8} \\
WideResNet-28-10 & 82.9 & 82.7 & \textbf{83.3} & 82.5 & \textbf{83.3} \\
\hline
\textbf{ImageNet} & & & & & \\
ResNet50 & \textbf{77.6} & \textbf{77.6} & \textbf{77.6} & 77.5 & 77.5 \\
\hline
\end{tabular}
\end{minipage}}
\label{tbl:smallscale}
\end{center}
\end{table}

\subsection{CIFAR evaluation}
\label{lbl:smallscale}

For both CIFAR datasets, we always make use of the following augmentations
: pad-and-crop by 4 pixels, as well as random horizontal flips and rectangular cutouts. Both WideResNet models were trained for 250 epochs with a learning rate of 0.1, batch size of 128, a Nesterov SGD optimizer with a weight decay of $5\cdot 10^{-4}$ and an annealing cosine learning rate decay. For WideResNets (WRN) and the ShakeShake models, we rely on the constant position model. Our filter training and measurement subset consists of 25600 and 512 samples respectively. Due to the different policy definition and application described in Sec.~\ref{lbl:cta}, the values we used for $l$ (number of nonzero states in a particle, similar to parameter $n$ in RA) and $m$ (augmentation magnitude) were different from RandAugment. For CIFAR-10, we use $l=3,\ m=3$ for the WRN-28-2 and ShakeShake models and $l=4,\ m=2$ for the WRN-28-10 model. Furthermore, the ShakeShake model was trained for 750 epochs with an initial learning rate of $0.03$ and a filter learning rate $\eta$ of $0.25$. For CIFAR-100, we set $l=2,\ m=2$ for the WRN-28-2 model and $l=4,\ m=6$ for the WRN-28-10 model. For the latter model, the constant position model was changed to a constant velocity model with $\mathbf{c}= -0.001\cdot[1\ \hdots\ 1]$, the initial nonzero states were initialized to $1.0$ instead of $0.5$ and the first 15 particles were set to 15 orthogonal unit vectors. The selected values for $l$ and $m$ were determined by a simple hyper-parameter optimization using grid search.

As presented in Table \ref{tbl:smallscale}, our ParticleAugment outperforms or matches the prior work for all experiments on CIFAR-10 and CIFAR-100. It should be noted that only 50 particles were sufficient to reach excellent performance. Furthermore, only a small computational overhead was introduced compared to an even faster scheme like DADA~\citep{li2020differentiable}. At the same time, we offer better model performance, compressing the optimization into a single training episode.

\subsection{ImageNet evaluation}

In our setup, we rely on a ResNet50 model and train it for 210 epochs. In addition to the augmentations applied by ParticleAugment, the default augmentations for ImageNet were always applied as well, i.e.~random crops, resizing to 224x224, random horizontal flips, and color jitter. The augmentation magnitude $m$ was set to 3. For the filter and training parameters, we chose identical values to the CIFAR-10 setup with the exception of an $\eta$ of 0.15 and the training subset size of 15\% of the training set. We employed distributed training on 4 GPUs and decreased the learning rate of ResNet to 0.05 and batch size to 64.

As shown in Table \ref{tbl:smallscale}, we reach similar performance to the related approaches. Note that we do not use any proxy models for optimization compared to AutoAugment~\citep{cubuk2019autoaugment} and no major hyper-parameter optimization compared to RandAugment~\citep{cubuk2019randaugment}, as we mainly rely on parameter values obtained from the CIFAR-10 experiment.

\subsection{Ablation study}
\label{lbl:ablation}

To investigate the influence of our algorithm parameters, we performed a number of ablation experiments. As the particle number $r$ has a direct impact on the performance, it is important to select the right parameters without requiring more computations. Another parameter of interest is the filter learning rate $\eta$. The ablation studies are performed with CIFAR-10 on WideResNet-28-2 with the same hyper-parameters as in Sec.~\ref{lbl:smallscale}.

\begin{table}
\begin{center}
\caption{We investigate the influence of filter parameters on the accuracy of a WideResNet-28-2 trained on CIFAR-10. Either the particle number or the filter learning rate were varied. The result is the accuracy in percentage, measured as mean and standard deviation from 4 runs.}
\vskip 0.15in
\begin{tabular}{lc}
\hline
Particle number  $r$ & Test Accuracy\\
\hline\hline
25 & 96.32$\pm$0.07\\
\hline
50 & 96.44$\pm$0.15\\
\hline
100 & 96.40$\pm$0.14\\
\hline
200& 96.29$\pm$0.09\\
\hline
Filter learning rate $\eta$ & Test Accuracy\\
\hline\hline
0.01 & 96.07$\pm$0.17\\
\hline
0.1 & 96.37$\pm$0.05\\
\hline
1 & 96.44$\pm$0.15\\
\hline
4 & 96.33$\pm$0.05\\
\hline
\end{tabular}
\label{tbl:abl}
\end{center}
\end{table}

\paragraph{Number of particles}

We tested between 25 and 200 particles where a higher particle number means that in the filter update more computation time is needed, as each particle needs to be tested on the measurement subset. The added benefit of more particles lies in the improved coverage of the augmentation space. As more policies are considered, there is a higher probability to find the optimal policies. The results are presented in Table \ref{tbl:abl}, where we observe that a higher number of particles doesn't necessarily increase the final  accuracy. When having a large number of particles, 
after the re-sampling step, mainly policies with the highest weights get sampled, thus effectively reducing the amount of unique policies.

\paragraph{Particle filter learning rate}
There is an optimal filter rate which maximizes the performance, as shown in Table \ref{tbl:abl}. If it is chosen to be too high, then the particles diverge very quickly and degenerate, lowering the performance. This is  similar to how the neural network training leads to performance degradation with overly big learning rates. If the learning rate is too small, then the filter does not update its policies quick enough to follow the network improvement, especially if the particle filter is not invoked every epoch. This can lead to sub-optimal performance. We also observed that the optimal filter learning rate may depend on the dataset and network, as can be seen from the differing hyperparameters for some experiments in Sec.~\ref{lbl:smallscale}.

\begin{table}
\begin{center}
\caption{
We compare our GPU hours (based on training on a NVIDIA GeForce 2080Ti) required for finding optimal policies with AutoAugment and DADA. The GPU hours required for these methods are obtained from AutoAugment \citep{cubuk2019autoaugment} and DADA \citep{li2020differentiable} publications.}
\vskip 0.15in
\begin{tabular}{lccc}
\hline
Setup & \multicolumn{3}{c}{GPU hours}\\
\hline\hline
Algorithm & AA & DADA & Ours\\
\hline
CIFAR-10 \& WRN-28-2 & 5000 & 0.1 & 3\\
\hline
CIFAR-100 \& WRN-28-10 & 5000\footnotemark & 0.2 & 6.6\\
\hline
ImageNet \& Resnet-50 & 15000 & 1.3 & 30\\
\hline
\end{tabular}
\label{tbl:abl2}
\end{center}
\end{table}

\paragraph{Process model choice}

As mentioned in Sec.~\ref{lbl:smallscale}, we made the observation that the constant velocity (CV) process model produces better results on CIFAR-100 than the constant position model for the bigger WideResNet (WRN). The motivation for the CV model came from observing favored policies during the last training epochs. To better represent these policies in the model, we introduced the velocity $c$ into the transition model, which is subtracted from the particles. We achieved a test accuracy of 96.38\%$\pm$0.10\% for the CV model and 96.44\%$\pm$0.15\% for the constant position (CP) model on the CIFAR-10 WRN-28-2 experiment over 4 test runs. Therefore, we can infer that the CV model, while helping us achieve a state of the art result on CIFAR-100, is not suitable for every model and dataset and needs to be chosen accordingly. In the case of CIFAR-100, using the CP model only yielded a best accuracy of 82.7\%.

\paragraph{Augmentation order randomization}

We made the assumption that the augmentation order of application does not have an impact on the overall performance. We tested a randomized application order of the 15 possible augmentations and achieved an accuracy of 96.32\%$\pm$0.11\% over 4 test runs. This shows that a different augmentation application order does not have a significant impact on the end result, considering also the standard deviation of the experiments. Additionally, the standard WRN-28-2 CIFAR-10 experiment was repeated 10 times to measure the standard deviation of the model accuracy, which resulted in a value of $96.43\pm0.11\%$ over 10 separate training episodes. We can therefore see that our approach produces consistent results regardless of the fact that it uses a probabilistic approach.

\paragraph{Training time vs.~model performance}

Table \ref{tbl:abl2} shows that while our approach takes more time than DADA, it is still very efficient compared to AutoAugment. At the same time, it reaches better model performance than both approaches. When changing the overall training to only 120 epochs, we still achieved 76.9\% accuracy on ResNet50 on the same compute instance. By changing the training schedule of a model, we can trade between an increase in accuracy and training time reduction.

\section{Conclusion}

We presented ParticleAugment to approximate the optimal image augmentation policies during the neural network training. Our algorithm relies on Monte Carlo sampling to explore the augmentation state space. We defined the policy space as a distribution function and relied on a particle filter to find the optimal policies. We proposed the measurement of the policy performance based on the loss function and also used it to re-weight the particles and for the policy update. In our evaluations, we showed improved results on standard benchmarks compared to prior work.
When comparing with the prior work, we demonstrate that our approach reaches a balance between the computational cost of policy search and the model performance. 

\section*{Acknowledgments}
Part of the work was supported by the LUKAS project (19A20004F), funded by German Federal Ministry for Economic Affairs and Energy (BMWi).





\bibliographystyle{model2-names}
\bibliography{egbib}



\end{document}